\documentclass[10pt,twocolumn,letterpaper]{article}

\usepackage{cvpr}              
\usepackage{multirow} 
\definecolor{cvprblue}{rgb}{0.21,0.49,0.74}
\usepackage[pagebackref,breaklinks,colorlinks,allcolors=cvprblue]{hyperref}

\def\paperID{5538} 
\def\confName{CVPR}
\def\confYear{2026}

\title{Requesting Expert Reasoning: Augmenting LLM Agents with Learned Collaborative Intervention}

\author{Zhiming Wang \quad Jinwei He \quad Feng~Lu \\
 State Key Laboratory of VR Technology and Systems, 
  School of CSE, Beihang University\\
{\tt\small \{zy2306418, sy2406412,lufeng\}@buaa.edu.cn}
}

\begin{document}
\maketitle
\begin{abstract}
Large Language Model (LLM) based agents excel at general reasoning but often fail in specialized domains where success hinges on long-tail knowledge absent from their training data. While human experts can provide this missing knowledge, their guidance is often unstructured and unreliable, making its direct integration into an agent's plan problematic. To address this, we introduce AHCE (Active Human-Augmented Challenge Engagement), a framework for on-demand Human-AI collaboration. At its core, the Human Feedback Module (HFM) employs a learned policy to treat the human expert as an interactive reasoning tool. Extensive experiments in Minecraft demonstrate the framework's effectiveness, increasing task success rates by 32\% on normal difficulty tasks and nearly 70\% on highly difficult tasks, all with minimal human intervention. Our work demonstrates that successfully augmenting agents requires learning how to request expert reasoning, moving beyond simple requests for help.

\end{abstract}    
\section{Introduction}
\label{sec:intro}
Large Language Models (LLMs) have demonstrated remarkable capabilities in general-purpose reasoning~\cite{qwen2025qwen25technicalreport, deepseekai2025deepseekr1incentivizingreasoningcapability, gemini, gpt4}. However, a significant challenge persists in enabling AI agents to solve domain-specific problems, where success often hinges on specialized expertise rather than common sense. This expertise, frequently derived from practical experience, is characterized by its rarity, contextual nuance, and the near impossibility of its comprehensive codification. Consequently, even the largest models trained on general corpora exhibit a critical failure of generalization when faced with situations that demand this type of long-tail, \textit{tacit} knowledge.

\begin{figure}
\begin{center}
\includegraphics[width=\linewidth]{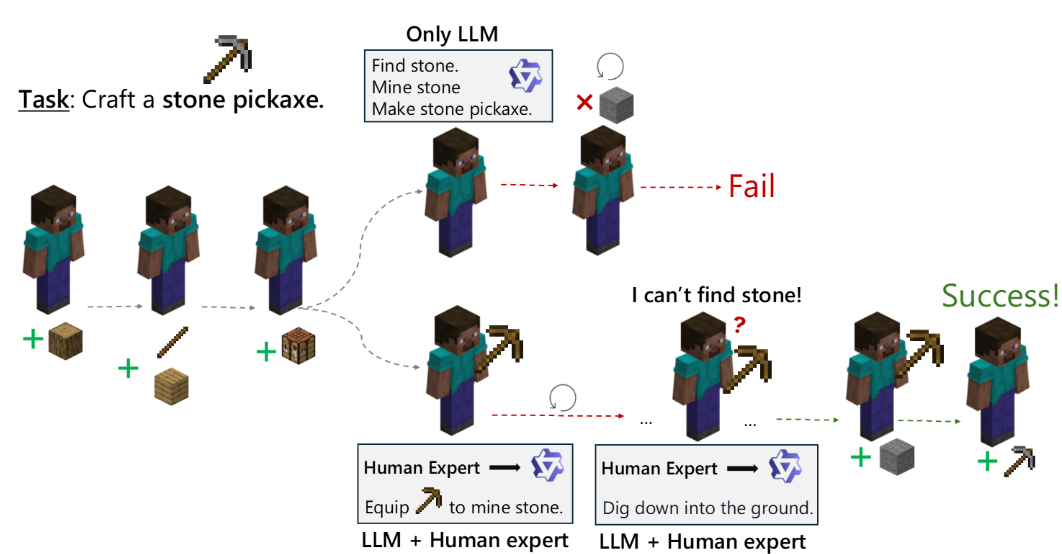}
\end{center}
\vspace{-3mm}

\caption{
Illustration of the multi-layered knowledge gap in domain-specific tasks. An agent's failure can stem from a missing \textbf{factual rule} (top path, trying to mine stone without a pickaxe) or a missing \textbf{strategic heuristic} (bottom path, not knowing to dig down for stone). Our approach (bottom path) leverages on-demand human expertise to address both types of failures, enabling successful task completion.}

\label{fig:idea}
\vspace{-3mm}
\end{figure}

This gap between general knowledge and domain-specific expertise is clearly illustrated in Minecraft~\cite{mp5, optimus_2, vpt, minedojo, minerl}, a widely-used testbed for AI agents. As shown in Figure~\ref{fig:idea}, an LLM-based agent attempting to craft a stone pickaxe can fail in two distinct ways. First, it might create a plan---find stone, mine stone, craft---that fails because it lacks a piece of explicit, factual knowledge: \textit{stone can only be mined with a pickaxe}. Even if this rule is learned, a second, more subtle failure can occur. After crafting a wooden pickaxe, the agent might wander the surface indefinitely, complaining "I can't find stone!" Here, it lacks practical, experiential knowledge that a human player has: \textit{stone is best found by digging downwards}. The first failure is a missing rule, while the second is a missing strategy. This shows that the knowledge gap has multiple layers, from simple facts to complex heuristics.

How can we bridge this multi-layered gap? Common approaches like using more data or external tools are insufficient. The data-centric approach of fine-tuning~\cite{optimus_2} could potentially teach the agent the first rule (a simple fact), but struggles with the second. It is nearly impossible to create a static dataset that covers all such practical strategies, like where and how to search for resources in different situations. Alternatively, using a static tool like a game wiki also falls short. A wiki can provide the crafting rule, but it cannot analyze the agent's specific context---wandering on the surface---and offer the strategic advice to "dig down". Since both data and static tools fail to address the deeper, strategic layer of knowledge, we conclude that access to a source capable of dynamic and contextual reasoning is required. At present, only a human expert can fill this role.

We therefore propose empowering the agent to collaborate with a human expert on-demand. As shown in Figure~\ref{fig:idea}, a human can resolve both types of failures with simple, targeted advice. However, the viability of such a human-in-the-loop system depends on two critical factors: the agent must maintain its autonomy by seeking help \textbf{only when truly necessary}, and it must be able to convert the unstructured human guidance into reliable action. Directly using raw feedback can be inefficient or counterproductive, as it may lead to new errors. The core research problem is therefore not just getting advice, but learning \textbf{when to ask} for it and \textbf{how to use} it effectively. This leads to our central goal: to design agents that learn to strategically request and leverage expert reasoning, enabling them to solve complex problems with minimal human oversight.

To operationalize this collaborative reasoning process, we introduce AHCE (Autonomous Human-in-the-loop Collaborative Enhancement), a modular framework for on-demand Human-AI collaboration. The process begins with the Problem Identification Module (PIM), which monitors the agent's execution history to autonomously detect critical impasses that signal the need for external guidance. Once an impasse is identified, the PIM activates our core innovation: the Human Feedback Module (HFM). Inspired by recent work on tool-augmented reasoning that integrates external knowledge retrieval into an LLM's thought process~\cite{research, searchr1, r1search}, we posit that the human expert can be \textbf{treated as} a unique, interactive tool. The HFM is therefore not a passive query mechanism, but an active synthesizer. It employs a learned policy to navigate a structured dialogue with the expert, probing for details and clarifying ambiguities. The outcome of this process is a robust, actionable plan, collaboratively refined by both the agent and the human. Finally, this synthesized plan is passed to the Query Execution Module (QEM), which \textbf{implements the corrected strategy}. This three-stage pipeline ensures that human guidance is sought only when necessary, and that this invaluable yet unstructured expertise is transformed into a reliable, machine-executable solution.

Extensive experiments demonstrate that our AHCE framework significantly improves agent performance on complex, open-world, process-dependent tasks. The framework achieves a 32\% increase in success rate on tasks of normal difficulty and a nearly 70\% increase on highly difficult tasks. Furthermore, our analysis explores the trade-off between agent autonomy and the frequency of human intervention.

In summary, our main contributions are:

\begin{itemize}
    \item We identify and address a critical yet often overlooked challenge in Human-in-the-Loop systems: the inherent unreliability and unstructured nature of raw human expertise, which can hinder agent performance.
    \item We propose AHCE, a novel framework that operationalizes a new paradigm of collaborative reasoning. Its core innovation is the Human Feedback Module (HFM), which employs a learned policy to treat the human expert as an interactive reasoning "tool," actively synthesizing unstructured guidance into a robust, executable plan.
    \item Through extensive experiments in Minecraft, we demonstrate that our method achieves substantial improvements in success rates across various difficulty levels. Crucially, these gains are realized with minimal human intervention, validating the efficacy and efficiency of our approach.
\end{itemize}

\section{Related Work}
\label{sec:RelatedWork}
\subsection{Reinforcement Learning with LLMs}
Reinforcement learning~\cite{reinforceLearning}, which aims to maximize the expected return of an agent’s policy through interactions with the environment, has emerged as a crucial technique for LLMs, from aligning with human values to enhancing reasoning capabilities. A significant development was Reinforcement Learning from Human Feedback (RLHF)~\cite{RLHF}, which uses Proximal Policy Optimization (PPO)~\cite{PPO} with reward models trained on human preferences. Several methods have since improved upon PPO, including Direct Preference Optimization (DPO)~\cite{DPO}, Simulated Preference Optimization (SimPO)~\cite{simpo}, and Group Relative Policy Optimization (GRPO)~\cite{GRPO}. Recently, several concurrent works have also begun to investigate reinforcement learning for enhancing LLM reasoning with tool use~\cite{searchr1, r1search, research}. In our work, we treat the human expert as a specialized tool and leverage reinforcement learning to improve the LLM's ability to reason in collaboration with this expert.

\subsection{Agents in Minecraft}

Early research on Minecraft agents employed techniques like hierarchical RL and reward shaping~\cite{cai2023open, jiang2024reinforcement, oh2017zero}.  Subsequent large-scale approaches involved pre-training on video data (VPT~\cite{vpt}), learning world models (DreamerV3~\cite{hafner2023mastering}), or using language-aligned representations for instruction following (MineCLIP~\cite{minedojo}, Steve-1~\cite{steve_1}). The current paradigm has shifted to using LLMs as zero-shot planners~\cite{voyager, wang2023describe, ghost_in_minecraft} and MLLMs for visual perception~\cite{mp5, optimus_2}. Despite this progress, these fully autonomous agents are fundamentally limited by the static knowledge of their underlying models. Our work deviates from this trend by enabling the agent to proactively seek minimal human guidance to overcome these knowledge gaps when facing unseen challenges.

\section{Preliminary Study: Limitations of a Zero-Shot Minecraft Agent}
\label{sec:Preli}

To ground our research, we first establish a baseline agent and conduct a preliminary study to diagnose its failure modes in open-world process-dependent tasks.

\begin{figure}
\begin{center}
\includegraphics[width=\linewidth]{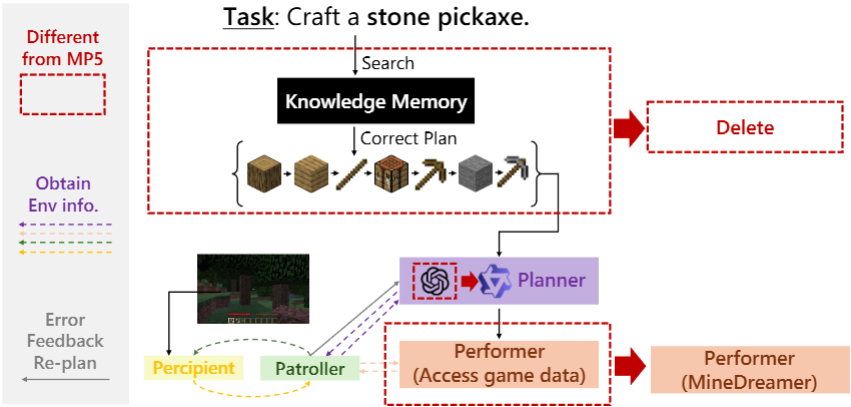}
\end{center}
\vspace{-3mm}

\caption{
The architecture of our MP5-core baseline, adapted from MP5 for a rigorous zero-shot evaluation. Key modifications include excising the Knowledge Memory to ensure true zero-shot planning, and replacing the original Performer with the vision-only MineDreamer~\cite{minedreamer} to operate without privileged game data.
}
\label{fig:mp5_vision}
\vspace{-3mm}
\end{figure}

\subsection{Baseline Implementation}
To rigorously evaluate our method, we required a baseline agent capable of operating in complex, open-ended worlds. While the MP5 framework~\cite{mp5} provides a conceptual blueprint for such an agent, its original implementation is unsuitable for a fair evaluation of zero-shot reasoning for two primary reasons. Therefore, we undertook a significant re-implementation effort to construct a baseline, which we term MP5-core, that addresses these fundamental issues.

As shown in Figure~\ref{fig:mp5_vision}. First, the framework suffers from knowledge contamination. Its memory module caches successful task decompositions from previous runs, providing the planner with solutions not derived from its intrinsic reasoning. This caching confounds any evaluation of its zero-shot performance. To address this, we remove the memory module, forcing all plans to be generated from the agent's core knowledge. Second, the original agent relies on privileged information, such as precise resource coordinates, which is unavailable in realistic scenarios and undermines the method's generalizability. To create a more realistic agent, we replace the original Performer with MineDreamer~\cite{minedreamer}, a vision-only controller that operates on raw perceptual input. Finally, for experimental consistency and reproducibility, we replace the planner's LLM from GPT-4~\cite{gpt4} to Qwen-plus~\cite{qwen}. The resulting baseline, which we term MP5-core, serves as a stringent testbed for evaluating agents under realistic, vision-only, zero-shot conditions.

\subsection{Preliminary Experiments}
Using our MP5-core baseline, we conducted a series of experiments with task settings detailed in Section 5. As shown in Table~\ref{tab:grouped_by_difficulty}, the agent achieves a perfect success rate on simple tasks. However, its performance degrades significantly as task difficulty increases. On difficult tasks, the success rate plummets to 10\%, falling to 0\% for chained tasks like mining iron ore.

\begin{figure}
\begin{center}
\includegraphics[width=\linewidth]{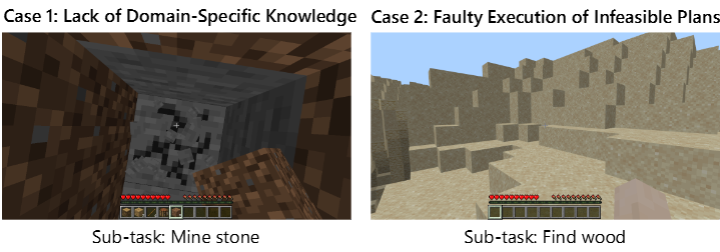}
\end{center}
\vspace{-3mm}

\caption{
Two primary failure modes for the autonomous agent. a planning failure from a domain-knowledge gap (left, mining stone without a pickaxe), and an execution failure from an environmental trap (right, searching for wood in a desert).
}
\label{fig:example_fail}
\vspace{-3mm}
\end{figure}

These failures stem from two primary sources. \textbf{Lack of Domain-Specific Knowledge}: The general LLM is unaware of Minecraft's fundamental rules. As shown in Figure~\ref{fig:example_fail} (left), it may instruct the agent to mine stone with its bare hands—an action that is logically plausible but physically ineffective in the game, which requires at least a wooden pickaxe. \textbf{Faulty Execution of Infeasible Plans}: The Performer struggles when the planner issues directives that are impossible in the current context. For instance, instructing the agent to gather wood in a desert where no trees exist (Figure~\ref{fig:example_fail}, right) leads the vision-based controller on a futile search, inevitably resulting in a timeout.

\begin{figure*}[h!]
\centering
\includegraphics[width=1\textwidth]{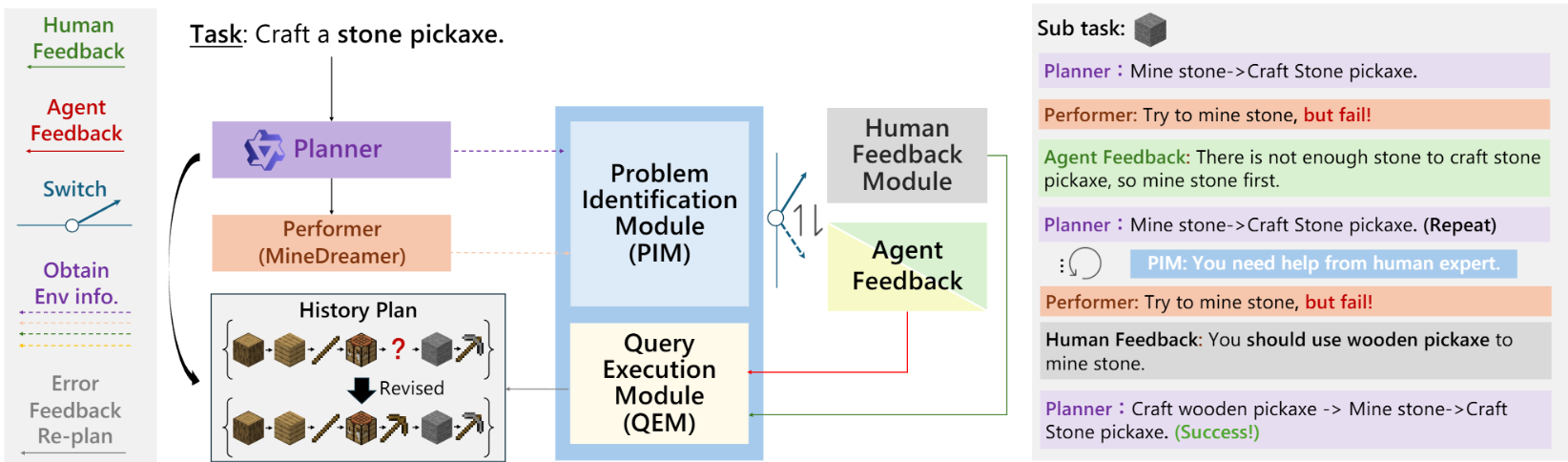}
\vspace{-0.5em}
\caption{An overview of our AHCE framework, which prioritizes autonomous self-correction (the red loop). When a critical impasse is detected, the system seamlessly transitions to solicit targeted human feedback (the green loop), enabling the agent to overcome knowledge gaps and achieve the task.}

\label{fig:AHCE}
\vspace{-1.0em}
\end{figure*}

\section{AHCE Framework}
\label{sec:Method}

\subsection{Overview}
\label{sec:overview}

As illustrated in Figure~\ref{fig:AHCE}, our AHCE framework enhances the MP5-core baseline by integrating three key modules: \textbf{the Problem Identification Module (PIM)}, \textbf{the Human Feedback Module (HFM)}, and \textbf{the Query Execution Module (QEM)}. The core principle of our design is to grant the agent maximum autonomy, enabling it to seek minimal human guidance only when it reaches a critical impasse. This approach is inspired by human problem-solving: attempting to find a solution first before seeking expert advice. To implement this principle, the agent first attempts to solve problems through several self-correction cycles. Only after these attempts fail does it conclude it is stuck and requests human assistance. This two-stage process is crucial for minimizing the frequency of human interventions and reducing the cognitive load on the expert.

The operational flow of AHCE begins when the agent receives a high-level task (e.g., "craft a stone pickaxe"), which the Planner, a zero-shot LLM, decomposes into sub-tasks. Upon failure, the agent enters a self-correction loop. If this fails, the PIM activates the human-in-the-loop protocol. At this juncture, the HFM, our central innovation, is invoked. The HFM is an LLM fine-tuned with reinforcement learning to master reasoning via tool-use. Our key insight is to conceptualize the human expert as a unique, interactive tool the LLM can learn to query. Consequently, rather than passively soliciting a complete solution, the HFM actively integrates the expert into its own iterative reasoning process. This interactive synthesis is critical: direct human advice can be unstructured or omit details, leading to unstable outcomes. By tasking the HFM to generate the final, structured corrective plan after incorporating the expert's insights, our approach ensures the resulting strategy is both robust and immediately actionable. Finally, the QEM executes this LLM-generated plan, allowing the agent to overcome the impasse and resume autonomous operation.

\begin{figure}
\begin{center}
\includegraphics[width=\linewidth]{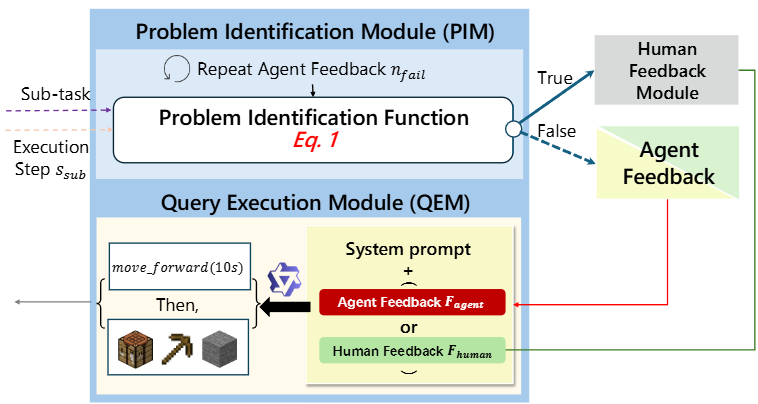}
\end{center}
\vspace{-3mm}

\caption{
The operational logic of the Problem Identification Module \textbf{(Top)} and Query Execution Modules \textbf{(Bottom)}.
}
\label{fig:pid_qem}
\vspace{-3mm}
\end{figure}

\subsection{Problem Identification Module}
The core function of the \textbf{PIM} is to enable the agent to autonomously recognize when it is truly stuck and requires external help. As depicted in Figure~\ref{fig:pid_qem} (top), this decision is governed by a effective mechanism based on two factors: sub-task execution timeout and cumulative failure count.

First, we define a sub-task failure. A sub-task is considered to have failed if its execution step count $s_{\text{sub}}$ exceeds a predefined maximum threshold $s_{\text{max}}$, indicating a timeout. Upon each failure, a counter for consecutive failures, $n_{\text{fail}}$, is incremented. The decision to seek help, $H$, is then determined by the following function:

\begin{equation}
    \label{eq:help_trigger}
    H(n_{\text{fail}}) = \begin{cases} 
    \text{True} & \text{if } n_{\text{fail}} > n_{\text{max}} \\
    \text{False} & \text{otherwise}
    \end{cases}
\end{equation}

where $n_{\text{max}}$ is a configurable hyperparameter representing the agent's degree of autonomy. If $H$ is False, the agent continues its self-correction loop; if True, the system switches to the Human Feedback Module (HFM). The value of $n_{\text{max}}$ directly controls the trade-off between task success and human burden. Based on our ablation study (Section 5.4), we determined that $\boldsymbol{n_{\text{max}}=3}$ strikes an effective balance between maximizing success rates on complex tasks and minimizing unnecessary interventions. Consequently, we adopt this value for all main experiments. As $n_{\text{max}} \to \infty$, our AHCE framework gracefully degrades to the fully autonomous MP5-core baseline.

\begin{figure*}[h!]
\begin{center}
\includegraphics[width=1\textwidth]{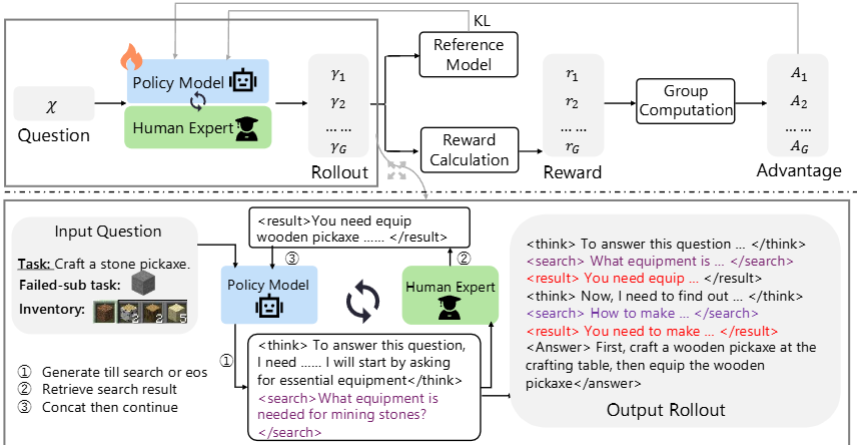}
\end{center}
\vspace{-3mm}

\caption{The overview of HFM. (a) The GRPO pipeline. (b) The detail of the rollout generation process.}
\label{fig:RL}
\vspace{-3mm}
\end{figure*}

\subsection{Human Feedback Module}
The Human Feedback Module (HFM) is designed to resolve critical impasses by learning an optimal policy for interacting with a human expert. At its core, the HFM reframes the problem of seeking help: instead of simply requesting a solution, the agent learns to use the human as an interactive tool to collaboratively synthesize a new plan. This approach is motivated by the observation that while expert guidance is invaluable, it can be unstructured or omit critical context, leading to unstable or incomplete plans if applied directly. By tasking an LLM to integrate this feedback into its own structured reasoning process, we can generate a final corrective plan that is both robust and immediately actionable.

To achieve this, we model the interaction as a sequential decision-making process and employ reinforcement learning to train the HFM, which is itself an LLM. The goal is to optimize the HFM's ability to generate a sequence of thoughts and queries that elicit and incorporate human knowledge.

\textbf{Group Relative Policy Optimization} Specifically, in this work, we use Group Relative Policy Optimization (GRPO) as the learning algorithm, which estimate the baseline from a group of rollouts instead of training a separate critic model in Proximal Policy Optimization (PPO). Given an existing policy $\pi_{\theta_{old}}$ and an reference policy $\pi_{\theta_{ref}}$, base on $G$ rollouts $\tau = \{y_i\}_{i=1}^{G} \sim \pi_{\theta_{\text{old}}}(\cdot|x)$ for each input $x \sim \mathcal{D}$, the objective of GRPO is to optimize the policy $\pi_{\theta}$ by maximizing the following objective. For brevity, we define the probability ratio $r_i(\theta) = \frac{\pi_\theta(y_i|x)}{\pi_{\theta_{\text{old}}}(y_i|x)}$. The objective is:

\begin{equation}
\label{eq:grpo}
\begin{aligned}
\mathcal{J}(\theta) = &\mathbb{E}_{\substack{x \sim \mathcal{D} \\ \{y_i\} \sim \pi_{\theta_{\text{old}}}}} \Bigg[ 
    \frac{1}{G} \sum_{i=1}^{G} \Bigg( \min \Bigg( 
        \frac{\pi_\theta(y_i|x)}{\pi_{\theta_{\text{old}}}(y_i|x)} A_i, \\
        &\quad \text{clip}\left( 
                \frac{\pi_\theta(y_i|x)}{\pi_{\theta_{\text{old}}}(y_i|x)}, 1-\epsilon, 1+\epsilon 
            \right) A_i 
    \Bigg) \\
    &- \beta \mathbb{D}_{\mathrm{KL}}(\pi_\theta \,\|\, \pi_{\theta_{\text{ref}}}) 
\Bigg] 
\end{aligned}
\end{equation}


where $A_i = \left( r_i - \text{mean}(\{r_j\}_{j=1}^{G}) \right) / \text{std}(\{r_j\}_{j=1}^{G})$ is the normalized advantage of the $i$-th rollout in current group, $\epsilon$ is the clipping ratio, and $\beta$ is the KL loss coefficient. Moreover, a KL divergenece penalty is added to the objective to prevent the policy from deviating too much from the original reference policy LLMs. The illustration of GRPO is shown in Figure~\ref{fig:RL}(a).

\paragraph{Interactive Reasoning with a Human-in-the-Loop.}
As depicted in Figure~\ref{fig:RL}(b), the HFM operates through an iterative generation process, mediated by special tags. The LLM generates its reasoning steps enclosed in \texttt{<think>} tags. When it needs external information, it formulates a query and encloses it within \texttt{<search>} tags. In our framework, this \texttt{<search>} action triggers an interaction with the human expert. The expert provides a textual response, which is then programmatically wrapped in \texttt{<result>} tags and appended to the LLM's current generation context. The LLM then continues its reasoning process from this enriched context, potentially issuing further queries until it has synthesized a complete, final plan encapsulated in \texttt{<Answer>} tags. This iterative loop allows the HFM to probe for details, clarify ambiguities, and fuse its own reasoning with the expert's knowledge.

\subsection{Query Execution Module}
\label{sec:query_execution}

Finally, the \textbf{QEM} (Figure~\ref{fig:pid_qem}, bottom part) is responsible for translating the high-level textual guidance of the HFM ($F_{\text{human}}$) into coherent executable strategies for both the Planner and the Performer.

\paragraph{Guiding the Planner.} The primary mechanism for course correction is to update the Planner's context. The HFM's advice is injected directly into the system prompt of the Planner's LLM. This acts as a form of dynamic, in-context learning, temporarily endowing the Planner with the specific domain knowledge it was lacking (e.g., "use a wooden pickaxe for stone"). 

\paragraph{Guiding the Performer.} For execution-level deadlocks, such as being stuck in an environment with no relevant resources (e.g., searching for wood in a desert), the HFM's advice can trigger low-level control policies. The \textbf{QEM} can parse phrases like "get out of the desert" and insert a pre-defined, procedural "escape maneuver" (e.g., \texttt{move\_forward(10s)}) into the plan. These primitive actions do not rely on privileged information and help the agent break out of unproductive local minima. By synergistically guiding both high-level planning and low-level execution, this module ensures that human expertise is effectively translated into task success.

\begin{table}[h!]
\centering
\caption{Categorization of tasks in Minecraft by level and complexity, defined by the approximate number of reasoning steps (i.e., minimum sub-objectives)  required for completion.}
\label{tab:task_levels}
\begin{tabular}{lcl}
\toprule
\textbf{Task Level} & \textbf{\# Reasoning Steps} & \textbf{Example Task} \\
\midrule
Easy \includegraphics[height=0.8em]{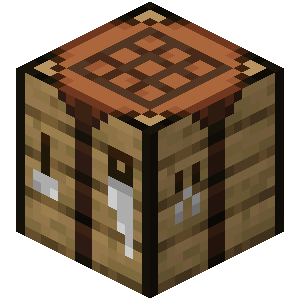}    & 1--3     & craft crafting table \includegraphics[height=0.8em]{figures/assets/minecraft_icons/crafting_table.png} \\
Normal \includegraphics[height=0.8em]{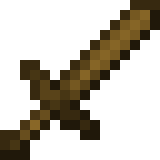}    & 4--5     & craft wooden sword \includegraphics[height=0.8em]{figures/assets/minecraft_icons/wooden_sword.png}  \\
Hard \includegraphics[height=0.8em]{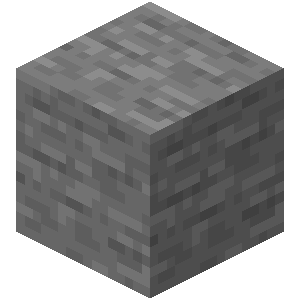}      & 6--9     & craft stone pickaxe \includegraphics[height=0.8em]{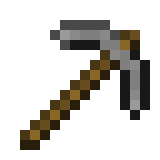} \\
\bottomrule
\end{tabular}
\end{table}
\section{Experiments}
\label{sec:Experiments}

\subsection{Experimental Setup}

\textbf{Environmental Setting.} We conduct our experiments in the MineDojo simulation environment~\cite{minedojo}. For each task trial, the agent is initialized in a procedurally generated world to ensure that our evaluation measures generalization across diverse environmental conditions. The HFM module in our framework is developed based on the Qwen-2.5-7B-Instruct and Qwen-2.5-32B-Instruct models~\cite{qwen2025qwen25technicalreport}. To isolate the HFM's ability to learn the skill of collaborative reasoning, rather than merely memorizing game-specific facts, we train it exclusively on the training set of MuSiQue~\cite{musique}, a multi-hop question-answering dataset. This approach trains the HFM to effectively use an external knowledge source (in our case, the human expert) without pre-exposing it to any Minecraft-specific knowledge.

\begin{table*}[t!]
\centering
\caption{Main experimental results comparing our AHCE methods against baselines across tasks of varying difficulty. ($^*$) indicates that this method is implemented by ourselves, for details see section~\ref{sec:Preli}.}
\label{tab:grouped_by_difficulty}

\small 
\setlength{\tabcolsep}{2pt} 

\newcommand{\mcell}[2][c]{\begin{tabular}[#1]{@{}c@{}}#2\end{tabular}}

\renewcommand{\arraystretch}{1.5} 

\begin{tabular}{l|cccc|cccc|cccc}
\hline
\multirow{3}{*}{Method} & 
\multicolumn{4}{c|}{Easy} & 
\multicolumn{4}{c|}{Medium} & 
\multicolumn{4}{c}{Hard} \\ \cline{2-13} 

& \mcell{ Success \\ Rate} & \mcell{Human \\ Time (s)} & \mcell{Total \\ Time (s)} &  \mcell{Human \\ Ratio} & 
  \mcell{Success \\ Rate} & \mcell{Human \\ Time (s)} & \mcell{Total \\ Time (s)} &  \mcell{Human \\ Ratio} & 
  \mcell{Success \\ Rate} & \mcell{Human \\ Time (s)} & \mcell{Total \\ Time (s)} & \mcell{ Human \\ Ratio} \\ \hline

MP5-core$^*$~\cite{mp5} & \textbf{100\%} & 0 & 251.4 & 0\% & 64\% & 0 & 373.0 & 0\% & 10\% & 0 & - & 0\% \\ 
AHCE-log & \textbf{100\%} & 0 &  251.4 & 0\% & 86\% & 81.0 & 499.5 & 16.2\%  & 68\% & 310.1 & 1513.7 & 20.5\% \\ 
\hline

\textbf{AHCE-Qwen-7B-Instruct} & \textbf{100\%} & 0 &  251.4 & 0\% & 94\% & 57.2 & 439.8 & 13.0\% & 78\% & 122.3 & 1325.8 & 9.2\% \\ 
\textbf{AHCE-Qwen-32B-Instruct} & \textbf{100\%} & 0 &  251.4 & 0\% & \textbf{96\%} & \textbf{32.7} & 433.4 & \textbf{7.5\%} & \textbf{82\%} & \textbf{79.4} & 1265.6 & \textbf{6.3\%} \\ \hline
\end{tabular}
\end{table*}

\textbf{Task Setting.} To assess the effectiveness of AHCE, we evaluate it on a suite of open-world, process-dependent tasks, a benchmark category proposed by MP5~\cite{mp5}. As detailed in Table~\ref{tab:task_levels}, these tasks consist of interdependent sub-task sequences where the failure of any single step results in the failure of the entire task. This task design is particularly suited for testing an agent's ability to overcome long-tail knowledge gaps in complex, continuous execution. Informed by the MP5 setup and our preliminary studies, we curated a benchmark of 15 distinct tasks, categorized by difficulty into Simple, Moderate, and Hard. A comprehensive list and detailed descriptions are provided in Appendix A.1.

\textbf{Baselines.} We compare AHCE against two critical baselines to isolate the impact of our proposed contributions: \textbf{MP5-core}: A fully autonomous agent driven by the base LLM, without any human-in-the-loop mechanism. This baseline measures the agent's zero-shot performance.
\textbf{AHCE-log}: An ablation of our framework where the intelligent HFM is removed. Instead, when an impasse occurs, the agent's historical action log is directly presented to the human expert for guidance. This baseline represents the naive "simple help-seeking" approach and allows us to quantify the value of the HFM's collaborative reasoning capability. 

Direct quantitative comparisons with other agents like VPT~\cite{vpt} or Voyager~\cite{voyager} are impractical due to fundamental differences in action spaces, perception models, and environmental assumptions, a challenge noted in the original MP5 study~\cite{mp5}. Our experiments therefore focus on a controlled internal comparison to rigorously test our central hypothesis.

\textbf{Evaluation Metrics.}
We recruited 10 human participants (7 male, 3 female, all with prior Minecraft experience) to serve as experts. Each participant conducted one full trial for all 15 tasks. Further details on the participant setup and briefing protocol are provided in the Supplementary Materials (Section A.2). We report the average performance across these trials using the following four key metrics:
\begin{itemize}
    \item \textbf{Average Success Rate (\%):} The percentage of tasks successfully completed for each difficulty category.
    \item \textbf{Average Human Interaction Time ($T_{\text{human}}$):} The wall-clock time measured from the moment the expert begins reviewing the agent's query to the moment they submit their guidance.
    \item \textbf{Average Total Execution Time ($T_{\text{total}}$):} The total wall-clock time from task initiation to completion. It comprises both agent-only execution time and human interaction time ($T_{\text{total}} = T_{\text{agent}} + T_{\text{human}}$).
    \item \textbf{Average Human Participation Ratio:} The fraction of the total execution time that required human involvement, calculated as $T_{\text{human}} / T_{\text{total}}$.
\end{itemize}

\subsection{Results of Open-ended Process-dependent Tasks}

This section evaluates the performance of our AHCE framework on open-world process-dependent tasks. Our analysis aims to quantify two key aspects: (1) the improvement in task success rates and (2) the cognitive load imposed on the human expert, as measured by interaction time.

\textbf{Human Assistance Significantly Improves Task Success Rates.} 
Table~\ref{tab:grouped_by_difficulty} summarizes our main findings. While all methods achieve a 100\% success rate on Easy tasks, the performance gap widens substantially as complexity increases. For Medium tasks, introducing human collaboration (AHCE-Qwen-32B-Instruct) improves the success rate from 64\% to 96\%. This effect is even more pronounced for Hard tasks, where the fully autonomous MP5-core baseline is nearly helpless (10\% success), while both AHCE variants boost performance to 78\% and beyond. These results confirm that on-demand human intervention is a highly effective strategy for overcoming the "long-tail knowledge" problem in complex tasks that are otherwise intractable for autonomous agents.

\paragraph{Collaborative Reasoning with HFM Maximizes Both Success and Efficiency.}
The central hypothesis of our work is that intelligent collaboration via the HFM is superior to naive help-seeking. The data strongly supports this claim. Comparing AHCE-log to our full AHCE framework reveals a clear trend: the HFM consistently improves performance across both success rate and human efficiency.
\begin{itemize}
\item \textbf{Higher Success Rate}: On Hard tasks, the best-performing AHCE-Qwen-32B model achieves an 82\% success rate, a significant 14-point improvement over the 68\% of AHCE-log. This suggests that the HFM's ability to synthesize a structured plan mitigates errors that can arise from applying raw, unstructured human advice.
\item \textbf{Lower Human Burden}: The HFM also dramatically reduces the cognitive load on the expert. For Hard tasks, the required human interaction time drops from 310.1 seconds for AHCE-log to just 79.4 seconds for AHCE-Qwen-32B—Instruct reduction of nearly 75\%. This efficiency gain is also reflected in the Human Ratio, which plummets from 20.5\% to a mere 6.3\%.
\end{itemize}
These findings validate that the HFM's collaborative reasoning is not just a marginal improvement; it fundamentally enhances the quality and efficiency of human-AI collaboration. The larger 32B model further amplifies these benefits, indicating that a more capable reasoning module can leverage human expertise more effectively.

\begin{figure*}[t!]
    \centering
    \includegraphics[width=1\textwidth]{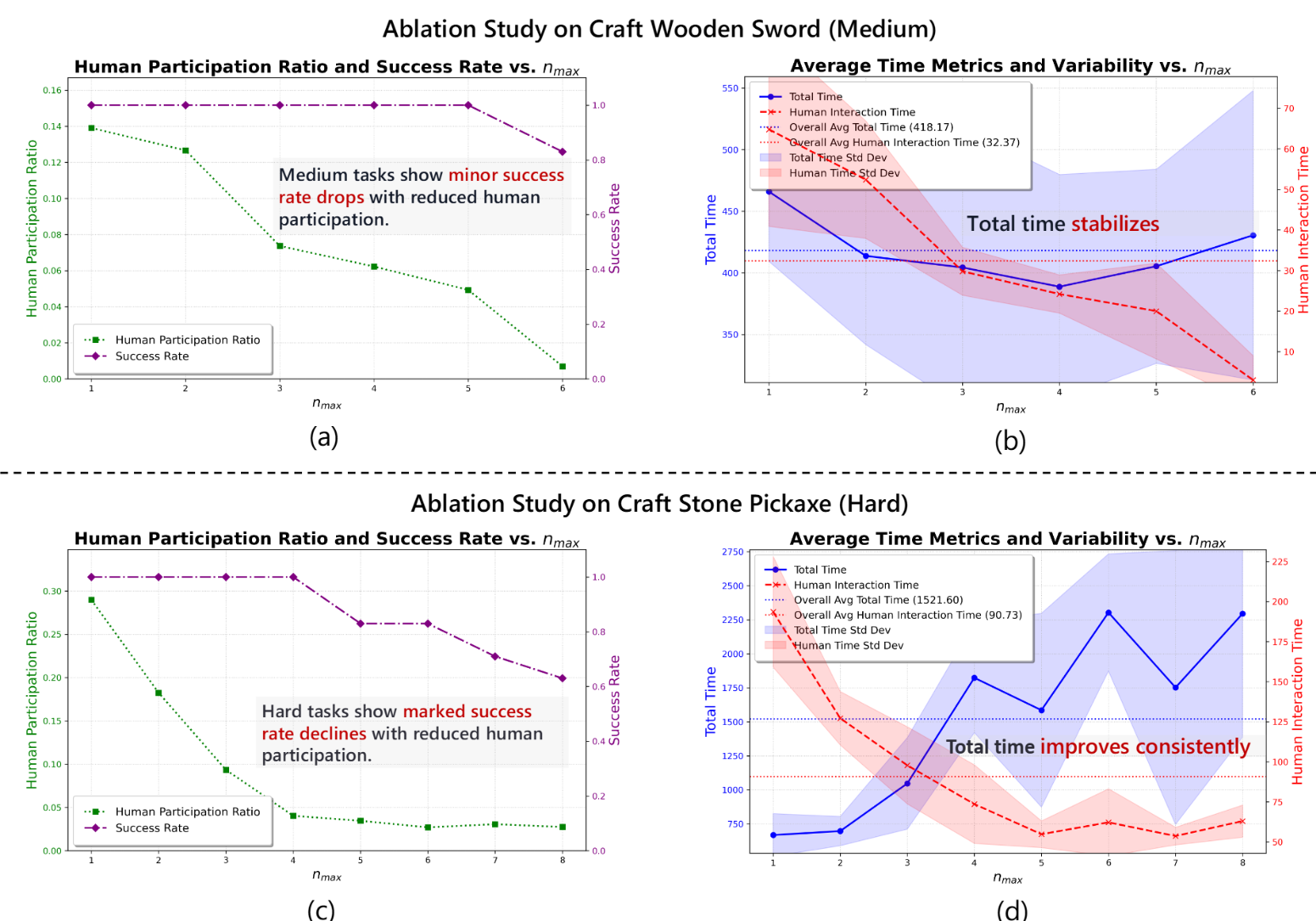}
    
    \vspace{-0.3cm}
    \caption{Ablation study on the failure threshold $n_{\text{max}}$, which governs the agent's autonomy before requesting human help. We evaluate the trade-off between task success rate and human interaction time on two tasks: medium difficulty (craft wooden sword) and hard difficulty (craft stone pickaxe). As $n_{\text{max}}$ increases, the agent acts more autonomously, reducing human effort but risking failure—especially on complex tasks.}
    \label{fig:ablation_study_n_max}
    \vspace{-0.4cm}
\end{figure*}

\subsection{Ablation Study}
To validate the critical role of the Problem Identification Module and quantify the trade-off between agent autonomy and task success, we conducted an ablation study on the failure threshold, $n_{max}$. This experiment was performed on two representative tasks: a moderately difficult task (craft wooden sword) and a highly complex task (craft stone pickaxe), using our AHCE-Qwen-32B-Instruct. As shown in Figure~\ref{fig:ablation_study_n_max}, we varied $n_{max}$ and measured its impact on both task success and human participation.

\textbf{Impact on Task Success and Human Effort.} Our findings reveal a clear, task-dependent relationship between agent autonomy ($n_{max}$), success rate, and human burden. For the simpler craft wooden sword task, the success rate remains highly robust, maintaining a perfect 100\% until $n_{max} > 5$ (Figure~\ref{fig:ablation_study_n_max} (a)). In stark contrast, for the more complex craft stone pickaxe task, the success rate plummets sharply for any $n_{max} > 3$ (Figure~\ref{fig:ablation_study_n_max} (c)). This confirms our hypothesis that for multi-stage tasks, excessive autonomy significantly increases the risk of the agent entering an irrecoverable state from which it cannot escape via self-correction alone.

Conversely, for both tasks, increasing $n_{max}$ predictably reduces the Human Participation Ratio. This is because a higher threshold allows the agent to complete more sub-tasks autonomously before needing help. The most significant reduction in human effort occurs at low values of $n_{max}$. For instance, on the craft stone pickaxe task, simply increasing $n_{max}$ from 1 to 3 cuts the human participation ratio by over half, from nearly 30\% down to 10\% (Figure~\ref{fig:ablation_study_n_max} (c)).

\textbf{Analysis of Task Completion and Intervention Time.} A deeper analysis of the time performance plots (Figure~\ref{fig:ablation_study_n_max} (b) and (d)) reveals further insights. For the complex craft stone pickaxe task, we observe a significant increase in the variance (light blue shaded area) of the Total Time as $n_{max}$ grows. This high variance suggests that with greater autonomy, the agent's behavior becomes more erratic; it may either get lucky or become trapped in long, unproductive failure loops, making its performance highly unpredictable.

Furthermore, the Human Interaction Time demonstrates a critical trend. While it decreases as $n_{max}$ increases, it does not approach zero. Instead, for the complex task, it plateaus at a non-zero minimum (approx. 50-60 seconds). This plateau represents the indispensable cognitive cost for a human to diagnose and correct the agent's core knowledge gaps—such as the need for a specific tool—which the agent cannot overcome on its own. This finding is crucial: for complex, process-dependent tasks, some level of human involvement is not just beneficial but essential for success. Attempting to eliminate this final piece of human guidance by setting an arbitrarily high $n_{max}$ would cause the agent's performance to degrade towards the near-zero success rate of the fully autonomous MP5-core baseline.

\section{Conclusion}
\label{sec:conclusion}


In this paper, we addressed the critical challenge of imbuing LLM-based agents with the specialized, long-tail knowledge required for complex, domain-specific tasks. We introduced a framework that empowers the agent to treat the human expert as an interactive reasoning tool. Through a learned policy, the agent learns not just to request help, but how to conduct a structured dialogue to reliably synthesize the expert's unstructured guidance into a robust plan. Our extensive experiments in Minecraft demonstrated the power of this approach, showing significant improvements in task success rates with minimal human intervention. Ultimately, our findings point toward a new paradigm for augmenting artificial intelligence. Instead of simply trying to pre-load agents with all possible knowledge, a more scalable strategy is to teach them the skill of collaborative reasoning. By learning how to request and leverage expert reasoning, agents can transcend the static boundaries of their training data and solve a new frontier of domain-specific problems.


{
    \small
    \bibliographystyle{ieeenat_fullname}
    \bibliography{main}

@String(CVPR= {IEEE Conf. Comput. Vis. Pattern Recog.})

@String(CVPR  = {CVPR})

@inproceedings{mp5,
  title={Mp5: A multi-modal open-ended embodied system in minecraft via active perception},
  author={Qin, Yiran and Zhou, Enshen and Liu, Qichang and Yin, Zhenfei and Sheng, Lu and Zhang, Ruimao and Qiao, Yu and Shao, Jing},
  booktitle={2024 IEEE/CVF Conference on Computer Vision and Pattern Recognition (CVPR)},
  pages={16307--16316},
  year={2024},
  organization={IEEE}
}

@inproceedings{optimus_2,
  title={Optimus-2: Multimodal minecraft agent with goal-observation-action conditioned policy},
  author={Li, Zaijing and Xie, Yuquan and Shao, Rui and Chen, Gongwei and Jiang, Dongmei and Nie, Liqiang},
  booktitle={Proceedings of the Computer Vision and Pattern Recognition Conference},
  pages={9039--9049},
  year={2025}
}

@article{vpt,
  title={Video pretraining (vpt): Learning to act by watching unlabeled online videos},
  author={Baker, Bowen and Akkaya, Ilge and Zhokov, Peter and Huizinga, Joost and Tang, Jie and Ecoffet, Adrien and Houghton, Brandon and Sampedro, Raul and Clune, Jeff},
  journal={Advances in Neural Information Processing Systems},
  volume={35},
  pages={24639--24654},
  year={2022}
}

@inproceedings{oh2017zero,
  title={Zero-shot task generalization with multi-task deep reinforcement learning},
  author={Oh, Junhyuk and Singh, Satinder and Lee, Honglak and Kohli, Pushmeet},
  booktitle={International Conference on Machine Learning},
  pages={2661--2670},
  year={2017},
  organization={PMLR}
}

@article{minedojo,
  title={Minedojo: Building open-ended embodied agents with internet-scale knowledge},
  author={Fan, Linxi and Wang, Guanzhi and Jiang, Yunfan and Mandlekar, Ajay and Yang, Yuncong and Zhu, Haoyi and Tang, Andrew and Huang, De-An and Zhu, Yuke and Anandkumar, Anima},
  journal={Advances in Neural Information Processing Systems},
  volume={35},
  pages={18343--18362},
  year={2022}
}

@article{minerl,
  title={Minerl: A large-scale dataset of minecraft demonstrations},
  author={Guss, William H and Houghton, Brandon and Topin, Nicholay and Wang, Phillip and Codel, Cayden and Veloso, Manuela and Salakhutdinov, Ruslan},
  journal={arXiv preprint arXiv:1907.13440},
  year={2019}
}

@article{gpt4,
  title={Gpt-4 technical report},
  author={Achiam, Josh and Adler, Steven and Agarwal, Sandhini and Ahmad, Lama and Akkaya, Ilge and Aleman, Florencia Leoni and Almeida, Diogo and Altenschmidt, Janko and Altman, Sam and Anadkat, Shyamal and others},
  journal={arXiv preprint arXiv:2303.08774},
  year={2023}
}

@article{gemini,
  title={Gemini: a family of highly capable multimodal models},
  author={Team, Gemini and Anil, Rohan and Borgeaud, Sebastian and Alayrac, Jean-Baptiste and Yu, Jiahui and Soricut, Radu and Schalkwyk, Johan and Dai, Andrew M and Hauth, Anja and Millican, Katie and others},
  journal={arXiv preprint arXiv:2312.11805},
  year={2023}
}

@article{voyager,
  title={Voyager: An open-ended embodied agent with large language models},
  author={Wang, Guanzhi and Xie, Yuqi and Jiang, Yunfan and Mandlekar, Ajay and Xiao, Chaowei and Zhu, Yuke and Fan, Linxi and Anandkumar, Anima},
  journal={arXiv preprint arXiv:2305.16291},
  year={2023}
}

@article{wang2023describe,
  title={Describe, explain, plan and select: Interactive planning with large language models enables open-world multi-task agents},
  author={Wang, Zihao and Cai, Shaofei and Chen, Guanzhou and Liu, Anji and Ma, Xiaojian and Liang, Yitao},
  journal={arXiv preprint arXiv:2302.01560},
  year={2023}
}

@article{ghost_in_minecraft,
  title={Ghost in the minecraft: Generally capable agents for open-world environments via large language models with text-based knowledge and memory},
  author={Zhu, Xizhou and Chen, Yuntao and Tian, Hao and Tao, Chenxin and Su, Weijie and Yang, Chenyu and Huang, Gao and Li, Bin and Lu, Lewei and Wang, Xiaogang and others},
  journal={arXiv preprint arXiv:2305.17144},
  year={2023}
}

@inproceedings{cai2023open,
  title={Open-world multi-task control through goal-aware representation learning and adaptive horizon prediction},
  author={Cai, Shaofei and Wang, Zihao and Ma, Xiaojian and Liu, Anji and Liang, Yitao},
  booktitle={Proceedings of the IEEE/CVF Conference on Computer Vision and Pattern Recognition},
  pages={13734--13744},
  year={2023}
}

@inproceedings{jiang2024reinforcement,
  title={Reinforcement learning friendly vision-language model for minecraft},
  author={Jiang, Haobin and Yue, Junpeng and Luo, Hao and Ding, Ziluo and Lu, Zongqing},
  booktitle={European Conference on Computer Vision},
  pages={1--17},
  year={2024},
  organization={Springer}
}

@article{hafner2023mastering,
  title={Mastering diverse domains through world models},
  author={Hafner, Danijar and Pasukonis, Jurgis and Ba, Jimmy and Lillicrap, Timothy},
  journal={arXiv preprint arXiv:2301.04104},
  year={2023}
}

@article{steve_1,
  title={Steve-1: A generative model for text-to-behavior in minecraft},
  author={Lifshitz, Shalev and Paster, Keiran and Chan, Harris and Ba, Jimmy and McIlraith, Sheila},
  journal={Advances in Neural Information Processing Systems},
  volume={36},
  pages={69900--69929},
  year={2023}
}

@article{minedreamer,
  title={Minedreamer: Learning to follow instructions via chain-of-imagination for simulated-world control},
  author={Zhou, Enshen and Qin, Yiran and Yin, Zhenfei and Huang, Yuzhou and Zhang, Ruimao and Sheng, Lu and Qiao, Yu and Shao, Jing},
  journal={arXiv preprint arXiv:2403.12037},
  year={2024}
}

@article{qwen,
  title={Qwen Technical Report},
  author={Jinze Bai and Shuai Bai and Yunfei Chu and Zeyu Cui and Kai Dang and Xiaodong Deng and Yang Fan and Wenbin Ge and Yu Han and Fei Huang and Binyuan Hui and Luo Ji and Mei Li and Junyang Lin and Runji Lin and Dayiheng Liu and Gao Liu and Chengqiang Lu and Keming Lu and Jianxin Ma and Rui Men and Xingzhang Ren and Xuancheng Ren and Chuanqi Tan and Sinan Tan and Jianhong Tu and Peng Wang and Shijie Wang and Wei Wang and Shengguang Wu and Benfeng Xu and Jin Xu and An Yang and Hao Yang and Jian Yang and Shusheng Yang and Yang Yao and Bowen Yu and Hongyi Yuan and Zheng Yuan and Jianwei Zhang and Xingxuan Zhang and Yichang Zhang and Zhenru Zhang and Chang Zhou and Jingren Zhou and Xiaohuan Zhou and Tianhang Zhu},
  journal={arXiv preprint arXiv:2309.16609},
  year={2023}
}

@article{reinforceLearning,
  title={Reinforcement learning},
  author={Sutton R S and Barto A G},
  journal={Journal of Cognitive Neuroscience},
  pages={126-134},
  year={1999}
}

@inproceedings{RLHF,
 author = {Ouyang, Long and Wu, Jeffrey and Jiang, Xu and Almeida, Diogo and Wainwright, Carroll and Mishkin, Pamela and Zhang, Chong and Agarwal, Sandhini and Slama, Katarina and Ray, Alex and Schulman, John and Hilton, Jacob and Kelton, Fraser and Miller, Luke and Simens, Maddie and Askell, Amanda and Welinder, Peter and Christiano, Paul F and Leike, Jan and Lowe, Ryan},
 booktitle = {Advances in Neural Information Processing Systems},
 editor = {S. Koyejo and S. Mohamed and A. Agarwal and D. Belgrave and K. Cho and A. Oh},
 pages = {27730--27744},
 publisher = {Curran Associates, Inc.},
 title = {Training language models to follow instructions with human feedback},
 url = {https://proceedings.neurips.cc/paper_files/paper/2022/file/b1efde53be364a73914f58805a001731-Paper-Conference.pdf},
 volume = {35},
 year = {2022}
}

@article{PPO,
  title={Proximal policy optimization algorithms},
  author={Schulman, John and Wolski, Filip and Dhariwal, Prafulla and Radford, Alec and Klimov, Oleg},
  journal={arXiv preprint arXiv:1707.06347},
  year={2017}
}

@article{DPO,
  title={Direct preference optimization: Your language model is secretly a reward model},
  author={Rafailov, Rafael and Sharma, Archit and Mitchell, Eric and Manning, Christopher D and Ermon, Stefano and Finn, Chelsea},
  journal={Advances in neural information processing systems},
  volume={36},
  pages={53728--53741},
  year={2023}
}

@article{simpo,
  title={Simpo: Simple preference optimization with a reference-free reward},
  author={Meng, Yu and Xia, Mengzhou and Chen, Danqi},
  journal={Advances in Neural Information Processing Systems},
  volume={37},
  pages={124198--124235},
  year={2024}
}

@article{GRPO,
  title={Deepseekmath: Pushing the limits of mathematical reasoning in open language models},
  author={Shao, Zhihong and Wang, Peiyi and Zhu, Qihao and Xu, Runxin and Song, Junxiao and Bi, Xiao and Zhang, Haowei and Zhang, Mingchuan and Li, YK and Wu, Yang and others},
  journal={arXiv preprint arXiv:2402.03300},
  year={2024}
}

@article{searchr1,
  title={Search-r1: Training llms to reason and leverage search engines with reinforcement learning},
  author={Jin, Bowen and Zeng, Hansi and Yue, Zhenrui and Yoon, Jinsung and Arik, Sercan and Wang, Dong and Zamani, Hamed and Han, Jiawei},
  journal={arXiv preprint arXiv:2503.09516},
  year={2025}
}

@article{r1search,
  title={R1-searcher: Incentivizing the search capability in llms via reinforcement learning},
  author={Song, Huatong and Jiang, Jinhao and Min, Yingqian and Chen, Jie and Chen, Zhipeng and Zhao, Wayne Xin and Fang, Lei and Wen, Ji-Rong},
  journal={arXiv preprint arXiv:2503.05592},
  year={2025}
}

@article{research,
  title={Learning to reason with search for llms via reinforcement learning},
  author={Chen, Mingyang and Sun, Linzhuang and Li, Tianpeng and Sun, Haoze and Zhou, Yijie and Zhu, Chenzheng and Wang, Haofen and Pan, Jeff Z and Zhang, Wen and Chen, Huajun and others},
  journal={arXiv preprint arXiv:2503.19470},
  year={2025}
}

@article{musique,
  title={MuSiQue: Multihop Questions via Single-hop Question Composition},
  author={Trivedi, Harsh and Balasubramanian, Niranjan and Khot, Tushar and Sabharwal, Ashish},
  journal={Transactions of the Association for Computational Linguistics},
  volume={10},
  pages={539--554},
  year={2022},
  publisher={MIT Press One Broadway, 12th Floor, Cambridge, Massachusetts 02142, USA~…}
}

@misc{qwen2025qwen25technicalreport,
      title={Qwen2.5 Technical Report}, 
      author={Qwen and : and An Yang and Baosong Yang and Beichen Zhang and Binyuan Hui and Bo Zheng and Bowen Yu and Chengyuan Li and Dayiheng Liu and Fei Huang and Haoran Wei and Huan Lin and Jian Yang and Jianhong Tu and Jianwei Zhang and Jianxin Yang and Jiaxi Yang and Jingren Zhou and Junyang Lin and Kai Dang and Keming Lu and Keqin Bao and Kexin Yang and Le Yu and Mei Li and Mingfeng Xue and Pei Zhang and Qin Zhu and Rui Men and Runji Lin and Tianhao Li and Tianyi Tang and Tingyu Xia and Xingzhang Ren and Xuancheng Ren and Yang Fan and Yang Su and Yichang Zhang and Yu Wan and Yuqiong Liu and Zeyu Cui and Zhenru Zhang and Zihan Qiu},
      year={2025},
      eprint={2412.15115},
      archivePrefix={arXiv},
      primaryClass={cs.CL},
      url={https://arxiv.org/abs/2412.15115}, 
}

@misc{deepseekai2025deepseekr1incentivizingreasoningcapability,
      title={DeepSeek-R1: Incentivizing Reasoning Capability in LLMs via Reinforcement Learning}, 
      author={DeepSeek-AI and Daya Guo and Dejian Yang and Haowei Zhang and Junxiao Song and Ruoyu Zhang and Runxin Xu and Qihao Zhu and Shirong Ma and Peiyi Wang and Xiao Bi and Xiaokang Zhang and Xingkai Yu and Yu Wu and Z. F. Wu and Zhibin Gou and Zhihong Shao and Zhuoshu Li and Ziyi Gao and Aixin Liu and Bing Xue and Bingxuan Wang and Bochao Wu and Bei Feng and Chengda Lu and Chenggang Zhao and Chengqi Deng and Chenyu Zhang and Chong Ruan and Damai Dai and Deli Chen and Dongjie Ji and Erhang Li and Fangyun Lin and Fucong Dai and Fuli Luo and Guangbo Hao and Guanting Chen and Guowei Li and H. Zhang and Han Bao and Hanwei Xu and Haocheng Wang and Honghui Ding and Huajian Xin and Huazuo Gao and Hui Qu and Hui Li and Jianzhong Guo and Jiashi Li and Jiawei Wang and Jingchang Chen and Jingyang Yuan and Junjie Qiu and Junlong Li and J. L. Cai and Jiaqi Ni and Jian Liang and Jin Chen and Kai Dong and Kai Hu and Kaige Gao and Kang Guan and Kexin Huang and Kuai Yu and Lean Wang and Lecong Zhang and Liang Zhao and Litong Wang and Liyue Zhang and Lei Xu and Leyi Xia and Mingchuan Zhang and Minghua Zhang and Minghui Tang and Meng Li and Miaojun Wang and Mingming Li and Ning Tian and Panpan Huang and Peng Zhang and Qiancheng Wang and Qinyu Chen and Qiushi Du and Ruiqi Ge and Ruisong Zhang and Ruizhe Pan and Runji Wang and R. J. Chen and R. L. Jin and Ruyi Chen and Shanghao Lu and Shangyan Zhou and Shanhuang Chen and Shengfeng Ye and Shiyu Wang and Shuiping Yu and Shunfeng Zhou and Shuting Pan and S. S. Li and Shuang Zhou and Shaoqing Wu and Shengfeng Ye and Tao Yun and Tian Pei and Tianyu Sun and T. Wang and Wangding Zeng and Wanjia Zhao and Wen Liu and Wenfeng Liang and Wenjun Gao and Wenqin Yu and Wentao Zhang and W. L. Xiao and Wei An and Xiaodong Liu and Xiaohan Wang and Xiaokang Chen and Xiaotao Nie and Xin Cheng and Xin Liu and Xin Xie and Xingchao Liu and Xinyu Yang and Xinyuan Li and Xuecheng Su and Xuheng Lin and X. Q. Li and Xiangyue Jin and Xiaojin Shen and Xiaosha Chen and Xiaowen Sun and Xiaoxiang Wang and Xinnan Song and Xinyi Zhou and Xianzu Wang and Xinxia Shan and Y. K. Li and Y. Q. Wang and Y. X. Wei and Yang Zhang and Yanhong Xu and Yao Li and Yao Zhao and Yaofeng Sun and Yaohui Wang and Yi Yu and Yichao Zhang and Yifan Shi and Yiliang Xiong and Ying He and Yishi Piao and Yisong Wang and Yixuan Tan and Yiyang Ma and Yiyuan Liu and Yongqiang Guo and Yuan Ou and Yuduan Wang and Yue Gong and Yuheng Zou and Yujia He and Yunfan Xiong and Yuxiang Luo and Yuxiang You and Yuxuan Liu and Yuyang Zhou and Y. X. Zhu and Yanhong Xu and Yanping Huang and Yaohui Li and Yi Zheng and Yuchen Zhu and Yunxian Ma and Ying Tang and Yukun Zha and Yuting Yan and Z. Z. Ren and Zehui Ren and Zhangli Sha and Zhe Fu and Zhean Xu and Zhenda Xie and Zhengyan Zhang and Zhewen Hao and Zhicheng Ma and Zhigang Yan and Zhiyu Wu and Zihui Gu and Zijia Zhu and Zijun Liu and Zilin Li and Ziwei Xie and Ziyang Song and Zizheng Pan and Zhen Huang and Zhipeng Xu and Zhongyu Zhang and Zhen Zhang},
      year={2025},
      eprint={2501.12948},
      archivePrefix={arXiv},
      primaryClass={cs.CL},
      url={https://arxiv.org/abs/2501.12948}, 
}
}


\end{document}